\documentclass{Interspeech}

\usepackage{tikz}
\usepackage{tcolorbox}
\usepackage{geometry}
\usepackage{times}
\usepackage{latexsym}
\usepackage{array}
\usepackage{pifont}
\usepackage{microtype}
\usepackage{float}
\usepackage{subcaption}
\usepackage{soul}

\usepackage{hyperref}

\usepackage{inconsolata}
\usepackage{makecell} 
\usepackage{graphicx}
\usepackage{booktabs} 
\usepackage{amsmath}
\usepackage{amssymb}
\interspeechcameraready

\title{The NaijaVoices Dataset: Cultivating Large-Scale, High-Quality, Culturally-Rich Speech Data for African Languages}

\author[affiliation={1,2,3,11}]{Chris}{Emezue}
\author[affiliation={1,2}]{The NaijaVoices}{Community}
\author[affiliation={4}]{Busayo}{Awobade}
\author[affiliation={1,2,5,10}]{Abraham Toluwase}{Owodunni}
\author[affiliation={2,7}]{Handel}{Emezue}
\author[affiliation={2,7}]{Gloria Monica Tobechukwu}{Emezue}
\author[affiliation={2,7}]{Nefertiti Nneoma}{Emezue}
\author[affiliation={6}]{Sewade}{Ogun}
\author[affiliation={8}]{Bunmi}{Akinremi}
\author[affiliation={3,12,13}]{David Ifeoluwa}{Adelani}
\author[affiliation={3,9,11,13}]{Chris}{Pal}

\affiliation{}{
Lanfrica, $^2$NaijaVoices, $^3$Mila - Quebec AI Institute,  $^4$MLCollective, $^5$The Ohio State University, $^6$INRIA, France, $^7$Alex Ekwueme Federal University Ndufu Alike Ikwo, Nigeria, \\$^8$Obafemi Awolowo University, Nigeria, $^9$Polytechnique Montreal, Canada, $^{10}$Newrun, \\$^{11}$University of Montreal, Canada, $^{12}$McGill University, Canada}{$^{13}$Canada CIFAR AI Chair}

\email{info@naijavoices.com}

\usepackage{comment}
\newcommand{\cmark}{\ding{51}}  
\newcommand{\xmark}{\ding{55}}  
\definecolor{naijaGreen}{HTML}{013800}

\begin{document}
\maketitle
\begin{abstract}

The development of high-performing, robust, and reliable speech technologies depends on large, high-quality datasets. However, African languages -- including our focus, Igbo, Hausa, and Yoruba -- remain under-represented due to insufficient data. Popular voice-enabled technologies do not support any of the 2000+ African languages, limiting accessibility for circa one billion people. While previous dataset efforts exist for the target languages, they lack the scale and diversity needed for robust speech models. To bridge this gap, we introduce the NaijaVoices dataset, a 1,800-hour speech-text dataset with 5,000+ speakers. We outline our unique data collection approach, analyze its acoustic diversity, and demonstrate its impact through finetuning experiments on automatic speech recognition, averagely achieving 75.86\% (Whisper), 52.06\% (MMS), and 42.33\% (XLSR) WER improvements. These results highlight NaijaVoices' potential to advance multilingual speech processing for African languages.

\end{abstract}

\section{Introduction \& Prior Work}

The importance of large, high-quality, and diverse training data in the performance of speech technologies cannot be overstated, as substantial datasets are essential for building high-performing speech processing models \cite{whisper, mms,longpre2023data,indicvoices,indicr,basetts}. While notable progress has been made in speech processing, African languages -- including our focus languages, Igbo, Hausa, and Yoruba  -- have largely been left behind \cite{nweya2022,ogunremi2023multilingual}. They are often categorized as `low-resource' \cite{orife2020masakhane,nekoto-etal-2020-participatory, joshi-etal-2020-state,adelani2022thousand} due to the lack of sufficient data needed to develop robust speech technologies. As a result, to this day, no single voice-enabled technology fully supports any of the 2,000+ African languages \cite{mozilla_howrwanda}.

Efforts have been made to create datasets for our focus languages, as analyzed in Table \ref{tab:comparison}. While these initiatives have been useful, their combined data quantity and diversity remain insufficient to build high-performing, robust speech systems comparable to those available for high-resource languages.

\begin{table}[ht!]
\centering
\resizebox{\linewidth}{!}{%
\begin{tabular}{p{3.5cm} >{\centering\arraybackslash}p{2.5cm} c c c}
\textbf{Dataset}&\parbox[c]{2.5cm}{\centering \textbf{Languages Supported}\\ \textbf{(Igbo, Hausa, Yoruba)}}& \textbf{\# Speakers} & \textbf{\# Hours} & \textbf{\# Sentences}  \\
\midrule
GlobalPhone \cite{schultz2013globalphone} & \xmark \cmark \xmark  & 82  & 6.6 & -\\
Lagos-NWU Yoruba \cite{olaleye2023yfacc} & \xmark \xmark \cmark  & 33   & 2.75   & 130 \\

IroyinSpeech \cite{ogunremi2023r}         & \xmark \xmark \cmark  & 80   & 42  &23K \\
YOR\`{U}LECT \cite{voicesungeard}        & \xmark \xmark \cmark  & 8  & 9.26   & 6K \\
BibleTTS \cite{meyer2022bibletts} & \xmark \cmark \cmark  & 1   & 119.9   & 50K \\
Common Voice (v.19) \cite{commonvoice} & \xmark \cmark \cmark & 170 &21&14K\\
IgboSynCorp \cite{nweya2022} & \cmark \xmark \xmark & 106 &38.8&- \\
FLEURS data set \cite{fleurs} & \cmark \cmark \cmark & 9 &54.50&11K \\
\addlinespace[2mm]
\textcolor{naijaGreen}{\textbf{NaijaVoices (Ours)}} & 
\textcolor{naijaGreen}{\cmark \cmark \cmark} & 
\textcolor{naijaGreen}{\textbf{5,455}} & 
\textcolor{naijaGreen}{\textbf{1,838.54}} & 
\textcolor{naijaGreen}{\textbf{645K}} \\

\bottomrule
\end{tabular}%
}
\caption{Comparative analysis of the NaijaVoices dataset with other speech datasets. The number of speaker, hours and sentences represent their combined sum over the languages supported.}
\label{tab:comparison}
\end{table}


To change this narrative, we have taken a bold step as a community by creating the NaijaVoices dataset, which comprises over 1,800 hours of diverse speech-text data from an unprecedented 5,000+ speakers. In this paper, we describe our dataset creation process, provide key statistics on its diversity and quality, and conduct finetuning experiments on automatic speech recognition to illustrate its potential for multilingual speech processing.

The NaijaVoices dataset is licensed under the CC BY-NC-SA 4.0 license and is accessible at \url{https://naijavoices.com/}.

\section{How NaijaVoices was created}


\begin{figure*}
    \centering
    \includegraphics[width=0.9\textwidth]{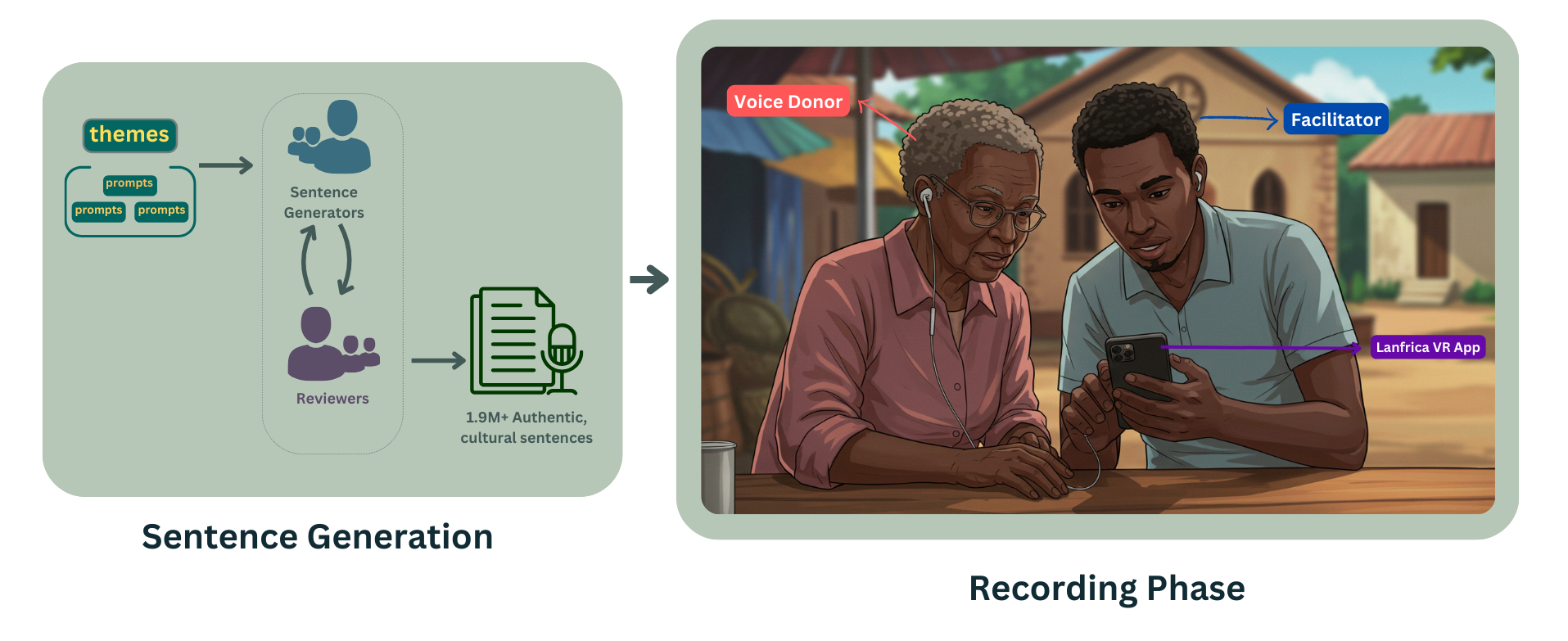}
\caption{Illustration of the creation of the NaijaVoices dataset. Beginning with the sentence generation phase where a group of language experts, called sentence generators, composed sentences based on themes/prompts provided. This generation phase was followed by a double review system, leading to over 1.9M texts which were recorded with our unique approach in the recording phase. Here, the `Facilitator' finds and closely guides the `Voice Donor' through the process of recording the generated sentences via the `VR App' for speech recording. Image depicting facilitator \& voice donor was generated using Google's ImageFX model with our prompt.}
    \label{fig:dataset-creation}
\end{figure*}

Our dataset creation process operates on our coined ethos of `data farming'\footnote{\url{https://lanfrica.com/blog/data-farming-in-the-global-south/}} which, in contrast to data mining which extracts and depletes data from providing communities, employs a reciprocal relationship (akin to farming) which ensures that providing communities are engaged in, empowered by, and mutually benefit from the data \st{collection} cultivation process. Figure \ref{fig:dataset-creation} illustrates our pipeline for creating the NaijaVoices dataset. The first challenge in building a novel dataset was sourcing sentences for recording. Conventional approaches, like web scraping, are cheap and easy to automate, but African languages are often under-represented and misrepresented online \cite{kreutzer2021quality,abdulmumin2022separating,penedo2024fineweb}. For instance, most texts available online for African languages are 
biblical texts \cite{orife2020masakhane,nekoto-etal-2020-participatory,meyer2022bibletts,jw300} or translations of Western-origin texts \cite{joshi-etal-2020-state,adelani2022thousand}. In an effort to obtain authentic, cultural texts, we created the sentences from scratch using our sentence generation approach.

\subsection{Sentence Generation} Language users and experts (both L1 and L2), called `Sentence Generators', manually composed new, culturally-aligned sentences of varying lengths and structures, guided by a total of 100 `themes', each of which had `prompts' -- descriptions and examples of the theme to elicit ideas in the generation phase. 
The sentence generation phase engaged 144 sentence generators (48 per language), and three language supervisors. Through a brainstorming session with language supervisors, a set of 100 themes/prompts, spanning a rich range of domains, was created for all three languages.  Each generator produced 48 sentences per theme, encompassing various structures (e.g., narrations, facts, commands, exclamations, questions-answers) and tenses (present, past, future) to ensure diversity. Success depended on carefully designed prompts, extensive training through workshops, detailed guidelines, and a thorough review system for quality control: each batch of generated sentences was reviewed by two reviewers and generators were required to correct any errors. For challenging cases -- such as new words/phrases without established native translations, discrepancies in diacritics, or other ambiguities -- language supervisors were consulted to address them collaboratively. Themes, in addition to being diverse, incorporated cultural elements (e.g., the theme `traditional counting systems' prompted generators to research traditional methods in their communities by consulting elders and old sources of information).

\subsection{Recording Phase} 
Many African speech dataset creation efforts \cite{nweya2022,voicesungeard,iroyin,ibrahim2021,wanjawa2022kencorpus} produce small outputs (tens of hours) due to conditions that, while offering much-needed supervision, limit scalability and diversity (such as requiring recorders to travel to a single recording booth for limited sessions one by one \cite{iroyin}). On the other hand, efforts to scale speech data creation to multiple recorders, typically through a bespoke speech recording app \cite{commonvoice} that allows multiple participants to record simultaneously, lead to subpar recordings mainly because of inadequate supervision: recorders are expected to navigate the app, comprehend recording guidelines (including what constitutes high-quality audio), and pronounce words correctly -- all on their own. This often leads to many erroneous recordings, due to numerous (and sometimes unprecedented) factors affecting the recorder's ability to produce high-quality recordings (technical background, understanding of the guidelines, to mention but a few).

With NaijaVoices, we figured out our unique approach to dataset creation that balances supervision and scalability, allowing us to capture thousands of speakers in a short time, while maintaining high-quality recordings. 
As depicted in Figure \ref{fig:dataset-creation}, in our recording phase, we hired `Facilitators’ -- individuals with technical skills, language expertise (particularly in pronunciation), mobility, deep understanding of their community and their reputation within it. These Facilitators underwent extensive training (about the project, the guidelines for obtaining high-quality recordings, the ethics of data collection, informed consent). Once trained, the Facilitators went into their communities in search of potential `voice donors'. They introduced the project to their communities, explained its purpose, how the data would be used, and answered questions posed. When someone agreed to donate their voice (becoming a `Voice Donor'), the Facilitator guided them through recording assigned sentences using our speech recording app (the `VR App'). During the recording process, Facilitators (who had linguistic expertise of the language) closely supervised the voice donors, ensuring words were pronounced correctly. 
Furthermore, we implemented offline and online variants of our approach, enabling greater flexibility, all of which allowed us to engage with over 5,000 voice donors (spanning all geopolitical zones of Nigeria) in an impressive three months.

A total of 1,917,686 sentences were recorded, with each sentence recorded twice and ensuring that no voice donor recorded the same sentence more than once. In addition to the Facilitators' inherent review process, we implemented a second review system where a group of reviewers evaluated a random subset of recordings from each Facilitator's voice donors. This feedback enabled us to monitor and assess the performance of the Facilitators.

The recording process involved two major categories, distinguished primarily by the number of sentences recorded per voice donor. The \textit{Phase 1} stage required each voice donor to record 240 sentences under the guidance of the Facilitator. We obtained a total of 1,269.21 hours recorded from approximately 5,220 voice donors in this phase. 
The \textit{Phase 2} stage, which focused on capturing longer audio recordings per speaker to support applications like text-to-speech, required each voice donor to record no fewer than 6,000 sentences. We obtained a total of 598.31 hours recorded from approximately 105 voice donors. 
\section{The NaijaVoices Dataset}

 The NaijaVoices dataset captures the essence of Nigerian culture through authentic, expert-generated, and culturally rich sentences, offering a level of originality rarely found in online texts \cite{kreutzer2021quality}. It features a wide range of speech patterns influenced by age, education levels, accents, and speaking styles -- from broken to formal speech, ethnic and dialectal influences. 
\subsection{Statistics}
An analysis of the dataset is displayed in Table \ref{table:dataset_statistics}, showing, for each language, the number of unique sentences, number of speakers, total hours, and descriptive statistics of the duration (in seconds). For the mean duration, the subscript represents the standard deviation. We observe that on average, the audios are around 3 - 4 seconds long.

\begin{table}[htb!]
\centering
\resizebox{\linewidth}{!}{
  \begin{tabular}{l c c c p{4cm}}
    \textbf{Language}  &  \textbf{\# Sentences} &           \textbf{\# Speakers} & \textbf{Total duration (h)} & \textbf{Duration statistics (seconds)} \\
    \hline
    \noalign{\vskip 2mm}
    Hausa   & 218K  & 1,879 & 605.27 & min: 0.06, max: 1,321.08, avg: 3.14$_{\pm2.11}$ \\ 
    Igbo    & 215K  & 1,808 & 626.33 & min: 0.06, max: 206.28,  avg: 3.89$_{\pm1.71}$ \\ 
    Yoruba  & 212K  & 1,768 & 606.94 & min: 0.06, max: 67.2,    avg: 3.56$_{\pm1.70}$ \\ 
    \noalign{\vskip 2mm}
    \hline
     \noalign{\vskip 2mm}
    Total   & 645K  & 5,455 & 1,838.54 & - \\ 
    \noalign{\vskip 2mm}
    \hline
  \end{tabular}
}
\caption{Detailed statistics on speech duration, unique sentence count, and speaker numbers for Hausa, Igbo, and Yoruba languages from the NaijaVoices dataset.}
\label{table:dataset_statistics}
\end{table}


     
All the unique speakers are assigned speaker IDs and anonymized, revealing only their age-range and gender in the publicly shared dataset. The NaijaVoices dataset consists of 58\% female and 42\% male speakers, with the majority falling within the 18–29 age range (65.9\%), followed by 6–17 years (25.7\%) and 30+ years (8.4\%), ensuring diverse demographic representation.

\subsection{Acoustic Diversity}

\begin{figure}[htb!]
    \centering
    \begin{subfigure}[t]{0.49\linewidth}
        \includegraphics[width=0.5\linewidth]{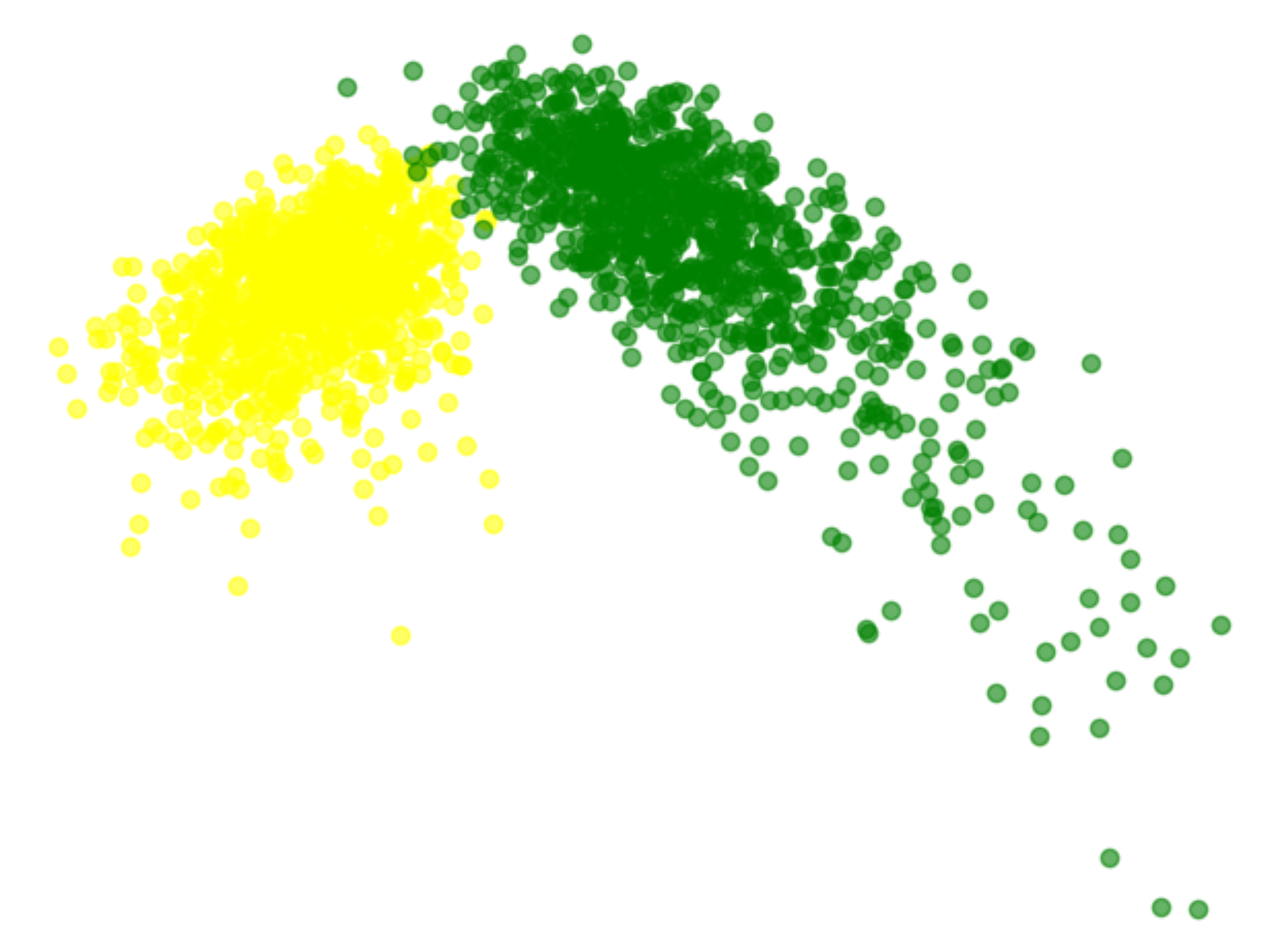}
        \caption{Hausa}
        \label{fig:acoustic-hausa}
    \end{subfigure}%
    \hfill%
    \begin{subfigure}[t]{0.49\linewidth}
        \includegraphics[width=0.5\linewidth]{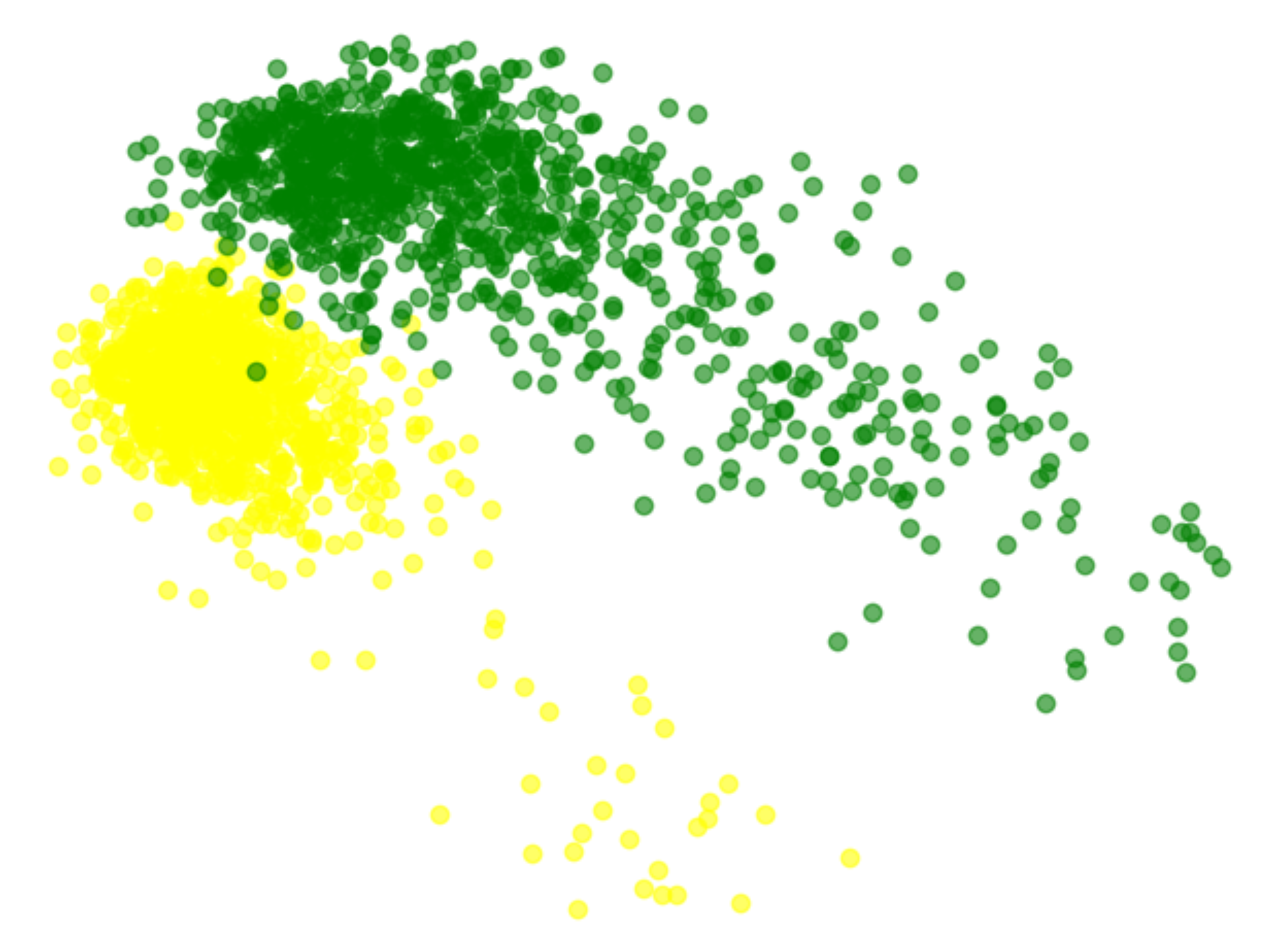}
        \caption{Yoruba}
        \label{fig:acoustic-yoruba}
    \end{subfigure}
    \caption{Acoustic diversity analysis of \textcolor{green}{NaijaVoices} and \textcolor{yellow}{Common Voice} datasets for Hausa and Yoruba languages. Using the unfinetuned MMS 1B model \cite{mms}, we extract 1,280-dimensional feature vectors (from the last layer of the encoder), reduce them to two dimensions with PCA \cite{pca}, and visualize them.}
    \label{fig:acoustic-diversity}
\end{figure}

The NaijaVoices dataset boasts a broad range of speaking styles from a diverse pool of speakers. To assess the acoustic diversity, we compare its acoustic feature representation to Common Voice v.19 \cite{commonvoice}, by sampling 1,000 Hausa and Yoruba recordings from each dataset (Igbo is not available in sufficient amount in Common Voice). Leveraging the unfinetuned MMS 1B model \cite{mms}, we extract 1,280-dimensional feature vectors (averaged over the sequence length of the audios) from the last layer of the encoder, reduce them to two dimensions with PCA \cite{pca}, and visualize them. As shown in Figure \ref{fig:acoustic-diversity}, NaijaVoices exhibits a broader dispersion, highlighting its coverage of diverse acoustic characteristics.

\subsection{Audio Quality}
\begin{figure}[h!]
    \centering
    \resizebox{\linewidth}{!}{
    \includegraphics[]{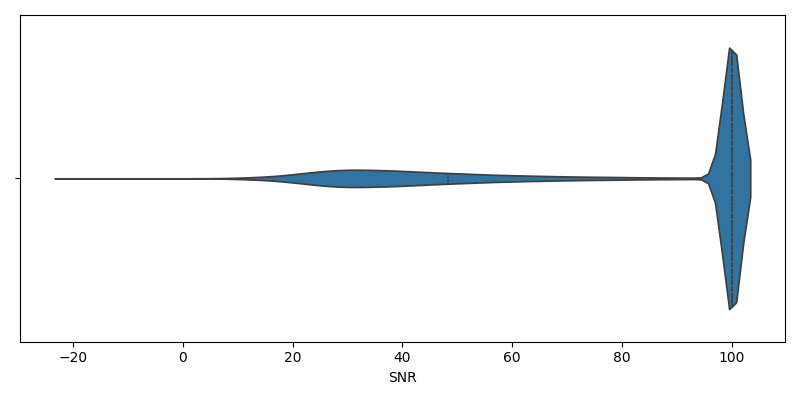}
    }
    \caption{Distribution of SNR values from audio samples in the NaijaVoices dataset, showing that the majority of audio samples lie within 100. According to \cite{wadasnr}, a value of 100, the highest possible value, means that the audio has a great proportion of silence w.r.t speech.}
    
    \label{fig:snr}
\end{figure}
To assess the audio quality, we compute the signal-to-noise ratio (SNR) for all audio samples using the WADA-SNR algorithm \cite{wadasnr}. As shown in Figure \ref{fig:snr}, approximately 65\% of the samples achieve the highest possible SNR score based on the WADA-SNR algorithm, while the remaining samples fall within the clean audio threshold (SNR 20–100) as defined by \cite{soapboxlabs_audio_quality_feature}. These results indicate that the NaijaVoices dataset consists of relatively clean audio recordings.

\begin{table*}[htb!]
\centering
\resizebox{0.8\textwidth}{!}{
\setlength{\tabcolsep}{4pt} 
\renewcommand{\arraystretch}{1.2} 
\begin{tabular}{lcccccc|c}
\textbf{Model} & \multicolumn{2}{c}{\textbf{Igbo}} & \multicolumn{2}{c}{\textbf{Yoruba}} & \multicolumn{2}{c}{\textbf{Hausa}} &  \\
 & NV Test & FLEURS & NV Test & FLEURS & NV Test & FLEURS & \textbf{Average} \\
\midrule
\textbf{XLSR (300M)} & & & & & & & \\
Unfinetuned & 100 & 100 & 100 & 100 & 100 & 100 & 100 \\
Monolingual Finetuned & \textbf{41.54} & \textbf{48.76} & 76.43 & 91.28 & 50.51 & 54.35 & 60.48 \\
Multilingual Finetuned & 52.44 & 58.42 & \textbf{65.14} & \textbf{69.21} & \textbf{49.73} & \textbf{51.10} & 57.67 \\
\midrule
\textbf{Whisper} & & & & & & & \\
Small Unfinetuned (244M) & 343.86 & 171.02 & 319.90 & 154.36 & 193.01 & 170.64 & 225.47 \\
Medium Unfinetuned (769M) & 262.68 & 224.42 & 196.80 & 204.84 & 199.55 & 139.17 & 204.58 \\
Large Unfinetuned (1.5B) & 102.60 & 101.49 & 106.23 & 103.56 & 106.58 & 144.33 & 110.80 \\
Small Finetuned (244M) & \textbf{53.67} & \textbf{74.65} & \textbf{58.10} & \textbf{79.83} & \underline{\textbf{21.00}} & \textbf{39.31} & 54.43 \\
\midrule
\textbf{MMS} & & & & & & & \\
Base Unfinetuned (300M) & 100 & 100 & 100 & 100 & 100 & 100 & 100 \\
Large Unfinetuned (1B) & 68.21 & \underline{\textbf{44.61}} & 73.14 & \underline{\textbf{53.56}} & 48.76 & \underline{\textbf{25.51}} & 52.30 \\
Monolingual Finetuned (300M) & \underline{\textbf{40.28}} & 47.31 & \underline{\textbf{51.65}} & 57.02 & \textbf{43.78} & 47.58 & \underline{\textbf{47.94}} \\
Multilingual Finetuned (300M) & 54.58 & 59.14 & 66.52 & 68.01 & 48.04 & 47.69 & 57.33 \\
\midrule
\textbf{SeamlessM4T v2} & & & & & & & \\
Large Unfinetuned (2.3B) & 94.73 & 96.90 & 67.36 & 83.5 & - & - & 85.62 \\

\bottomrule
\end{tabular}
}
\caption{Percentage Word Error Rates ($\downarrow$ WER\%) for different models across languages and test sets. \textbf{Scores in bold} indicate the best performance per language within each model family, while \underline{underlined scores} represent the best performance per language across all models. Please note that the WER metric is unbounded and values above 100\% are possible, indicating very poor performance \cite{huggingface}}.
\label{tab:eval-results}
\end{table*}

\section{Automatic Speech Recognition}

We perform automatic speech recognition (ASR) experiments to demonstrate the potential of the NaijaVoices dataset for multilingual speech research. Concretely, we finetune three selected ASR models on our dataset and evaluate them on both our test set (NV Test) and the FLEURS test set \cite{fleurs}.

\subsection{Experimental Setup}

\textbf{Model.}
For our ASR experiment, we select the following models for finetuning: Whisper small (244M), medium (769M), and large (1.5B) \cite{whisper}; the base XLSR model (300M) \cite{xlsr}; the base (300M), large (1B) MMS \cite{mms} and large (2.3B) Seamless M4T v2 \cite{seamless2023}. Additionally, we explore both monolingual and multilingual finetuning for XLSR and MMS to assess their adaptability across multiple languages. For Whisper, we finetune the small (244M) model while also evaluating its medium and large unfinetuned versions. The MMS model is tested in its base (300M) and large (1B) configurations, alongside its monolingual and multilingual finetuned variants.  Except for MMS pretrained on all three languages, Whisper and XLSR were originally pretrained on Yoruba and Hausa but not Igbo, while Seamless M4T v2 Large was pretrained on Igbo and Yoruba, excluding Hausa. Due to computational constraints, we finetune only the small and medium models, using their large variants as benchmarks to compare the performance of finetuned small models (on our dataset) against state-of-the-art large ASR models.

\textbf{Training Configurations.}
All finetuned models were trained for five epochs, and the base models were fully unfrozen during training. Following \cite{whisper}, the Whisper model was finetuned using a learning rate of 1e-5, with training and validation batch sizes set to 64 and a gradient accumulation step of 1. The pre-trained Whisper tokenizer was used for Yoruba and Hausa, while for Igbo, the English tokenizer was employed, as the Whisper tokenizer was not originally trained for Igbo. The XLSR and MMS model were finetuned under similar hardware conditions for 5 epochs, with a gradient accumulation step of 16. Training and validation batch sizes were set to 16, and, where needed, a custom tokenizer was constructed from the NaijaVoices dataset.

\textbf{Data Split.}
To ensure balanced representation across language, age bracket, and gender, we employed stratified sampling (by language, age bracket, and gender) in our dataset split. 99.7\% of the NaijaVoices dataset was allocated to the training set, with the remaining 0.3\% evenly divided into dev and test sets. To construct the NV Test set, we additionally selected all samples from 20 unseen speakers (those with the fewest samples), ensuring no overlap with the training and dev sets. This resulted in a total split size of 1832.73 hours for training, 3.02 hours for dev, and 3.34 hours for NV Test set across all languages.



\subsection{Results \& Discussion}

Table \ref{tab:eval-results} reports WER scores for both finetuned and unfinetuned models. Baseline models performed poorly, with WER reaching 100\% for XLSR and MMS, while Whisper’s smaller models also showed high error rates. Finetuning with NaijaVoices significantly improved performance, with multilingual finetuning benefiting Yoruba and Hausa, while monolingual finetuning was more effective for Igbo. Whisper’s finetuned small (244M) model showed the best results in Hausa but struggled with Yoruba and Igbo. The MMS finetuned (300M) model outperformed its unfinetuned counterpart, and the 1B-all unfinetuned model performed better on FLEURS but struggled with NV Test. Overall, the MMS monolingual finetuned model achieved the best average performance. Interestingly, the relatively smaller models, finetuned on our dataset, outperformed (on both the NV Test and the FLEURS test sets) Seamless M4T v2, a large state-of-the-art ASR model, attesting to the impact of large, high-quality training datasets \cite{ogueji-etal-2021-small,whang2021data,mishra2022proposal,10764812}.

These results affirm that our NaijaVoices dataset is a valuable dataset for improving ASR models, significantly reducing WER and underscoring the need for large-scale, high-quality language resources for African languages.

\section{Conclusion}
With the NaijaVoices dataset, we demonstrated the possibility of cultivating speech data for African languages at an unprecedented large scale (in terms of number of hours and speakers). Built on the principles of `data farming', our approach fosters a symbiotic relationship with language communities. Our dataset demonstrates unparalleled acoustic scale and diversity, surpassing existing resources for Igbo, Hausa, and Yoruba languages. We validated its impact through ASR finetuning experiments, achieving significant WER reductions across multiple models (on average we attain 75.86\% (Whisper), 52.06\% (MMS), and 42.33\% (XLSR) WER improvements) with just 5 epochs of finetuning. Moving forward, we are optimistic that NaijaVoices will serve as a catalyst for pushing the boundaries of multilingual speech processing for Igbo, Hausa, and Yoruba languages, finally lifting them from the `low-resource' status in terms of speech data, and ultimately driving greater inclusivity in voice-enabled technologies. Future work involves applying the principles of data farming, as well as our unique recording framework, to speech data cultivation for more African languages. 




\section{Acknowledgement}
We express our profound gratitude to the entire NaijaVoices community for making this dataset possible, and the Lacuna Fund for funding the creation of the NaijaVoices dataset. Finally we acknowledge the support of the IVADO and the Canada First Research Excellence Fund (CFREF) / Apogée Funds, the Canada CIFAR AI Chairs Program, as well as Mila - Quebec AI Institute and MLCollective for compute resources, all of which have been invaluable to our research.

\clearpage

\bibliographystyle{IEEEtran}
\bibliography{anthology}

\begin{thebibliography}{10}
\providecommand{\url}[1]{#1}
\csname url@samestyle\endcsname
\providecommand{\newblock}{\relax}
\providecommand{\bibinfo}[2]{#2}
\providecommand{\BIBentrySTDinterwordspacing}{\spaceskip=0pt\relax}
\providecommand{\BIBentryALTinterwordstretchfactor}{4}
\providecommand{\BIBentryALTinterwordspacing}{\spaceskip=\fontdimen2\font plus
\BIBentryALTinterwordstretchfactor\fontdimen3\font minus \fontdimen4\font\relax}
\providecommand{\BIBforeignlanguage}[2]{{%
\expandafter\ifx\csname l@#1\endcsname\relax
\typeout{** WARNING: IEEEtran.bst: No hyphenation pattern has been}%
\typeout{** loaded for the language `#1'. Using the pattern for}%
\typeout{** the default language instead.}%
\else
\language=\csname l@#1\endcsname
\fi
#2}}
\providecommand{\BIBdecl}{\relax}
\BIBdecl

\bibitem{whisper}
A.~Radford, J.~W. Kim, T.~Xu, G.~Brockman, C.~McLeavey, and I.~Sutskever, ``Robust speech recognition via large-scale weak supervision,'' \emph{ICML}, 2022.

\bibitem{mms}
V.~Pratap, A.~Tjandra, B.~Shi, P.~Tomasello, A.~Babu \emph{et~al.}, ``Scaling speech technology to 1,000+ languages,'' \emph{JMLR}, vol.~25, no.~97, pp. 1--52, 2024.

\bibitem{longpre2023data}
S.~Longpre, R.~Mahari, A.~Chen, N.~Obeng-Marnu, D.~Sileo \emph{et~al.}, ``The data provenance initiative: A large scale audit of dataset licensing \& attribution in {AI},'' \emph{arXiv preprint arXiv: 2310.16787}, 2023.

\bibitem{indicvoices}
T.~Javed, J.~Nawale, E.~George, S.~Joshi \emph{et~al.}, ``{I}ndic{V}oices: Towards building an inclusive multilingual speech dataset for {I}ndian languages,'' in \emph{Findings of ACL 2024}, aug 2024, pp. 10\,740--10\,782.

\bibitem{indicr}
A.~Sankar, S.~Anand, P.~S. Varadhan, S.~Thomas \emph{et~al.}, ``{IndicVoices-R}: Unlocking a massive multilingual multi-speaker speech corpus for scaling indian {TTS},'' \emph{NeurIPS DB}, 2024.

\bibitem{basetts}
M.~Łajszczak, G.~Cámbara, Y.~Li, F.~Beyhan \emph{et~al.}, ``{BASE TTS}: Lessons from building a billion-parameter text-to-speech model on 100k hours of data,'' \emph{arXiv preprint arXiv: 2402.08093}, 2024.

\bibitem{nweya2022}
\BIBentryALTinterwordspacing
G.~O. Nweya, A.~S. Oluwole, E.~F. Onwuegbuzia, S.~O. Ejinwa \emph{et~al.}, ``Replication data for {Igbo Natural Language Processing Tasks I},'' 2022. [Online]. Available: \url{https://doi.org/10.7910/DVN/RXBNCZ}
\BIBentrySTDinterwordspacing

\bibitem{ogunremi2023multilingual}
T.~{\`O}g{\'u}nr{\`e}m{\'i}, C.~D. Manning, and D.~Jurafsky, ``Multilingual self-supervised speech representations improve the speech recognition of low-resource {African} languages with codeswitching,'' \emph{CALCS}, 2023.

\bibitem{orife2020masakhane}
I.~Orife, J.~Kreutzer, B.~Sibanda, D.~Whitenack \emph{et~al.}, ``{Masakhane} - machine translation for {Africa},'' \emph{arXiv preprint arXiv: 2003.11529}, 2020.

\bibitem{nekoto-etal-2020-participatory}
W.~Nekoto, V.~Marivate, T.~Matsila, T.~Fasubaa \emph{et~al.}, ``Participatory research for low-resourced machine translation: A case study in {A}frican languages,'' in \emph{Findings of EMNLP 2020}, Online, nov 2020, pp. 2144--2160.

\bibitem{joshi-etal-2020-state}
P.~Joshi, S.~Santy, A.~Budhiraja, K.~Bali, and M.~Choudhury, ``The state and fate of linguistic diversity and inclusion in the {NLP} world,'' in \emph{Proceedings of ACL}, Online, jul 2020, pp. 6282--6293.

\bibitem{adelani2022thousand}
D.~I. Adelani, J.~O. Alabi, A.~Fan, J.~Kreutzer, X.~Shen, M.~Reid, D.~Ruiter, D.~Klakow \emph{et~al.}, ``A few thousand translations go a long way! {Leveraging} pre-trained models for {African} news translation,'' \emph{NAACL}, 2022.

\bibitem{mozilla_howrwanda}
\BIBentryALTinterwordspacing
R.~Muhire. (2020) How {Rwanda} is making voice tech more open. [Online]. Available: \url{https://foundation.mozilla.org/en/blog/how-rwanda-making-voice-tech-more-open/}
\BIBentrySTDinterwordspacing

\bibitem{schultz2013globalphone}
T.~Schultz, N.~T. Vu, and T.~Schlippe, ``{GlobalPhone}: A multilingual text \& speech database in 20 languages,'' in \emph{ICASSP}.\hskip 1em plus 0.5em minus 0.4em\relax IEEE, 2013, pp. 8126--8130.

\bibitem{olaleye2023yfacc}
K.~Olaleye, D.~Onea{\c{t}}{\u{a}}, and H.~Kamper, ``{YFACC}: A {Yor{\`u}b{\'a}} speech--image dataset for cross-lingual keyword localisation through visual grounding,'' in \emph{SLT 2022}.\hskip 1em plus 0.5em minus 0.4em\relax IEEE, 2023, pp. 731--738.

\bibitem{ogunremi2023r}
\BIBentryALTinterwordspacing
T.~Ògúnrèmí, K.~Túbosún, A.~Anuoluwapo, I.~Orife, and D.~I. Adelani, ``Ìròyìnspeech: A multi-purpose {Yorùbá} speech corpus,'' \emph{Proceedings of LREC}, 2023. [Online]. Available: \url{https://arxiv.org/abs/2307.16071v2}
\BIBentrySTDinterwordspacing

\bibitem{voicesungeard}
O.~Ahia, A.~Aremu, D.~Abagyan, H.~Gonen, D.~I. Adelani \emph{et~al.}, ``{Voices Unheard}: {NLP} resources and models for {Yorùbá} regional dialects,'' \emph{Findings of EMNLP}, 2024.

\bibitem{meyer2022bibletts}
J.~Meyer, D.~Adelani, E.~Casanova, A.~Öktem \emph{et~al.}, ``{BibleTTS}: a large, high-fidelity, multilingual, and uniquely {African} speech corpus,'' in \emph{INTERSPEECH}, 2022, pp. 2383--2387.

\bibitem{commonvoice}
\BIBentryALTinterwordspacing
R.~Ardila, M.~Branson, K.~Davis, M.~Kohler, J.~Meyer \emph{et~al.}, ``{Common Voice}: {A} massively-multilingual speech corpus,'' in \emph{Proceedings of LREC 2020}.\hskip 1em plus 0.5em minus 0.4em\relax European Language Resources Association, 2020, pp. 4218--4222. [Online]. Available: \url{https://aclanthology.org/2020.lrec-1.520/}
\BIBentrySTDinterwordspacing

\bibitem{fleurs}
A.~Conneau, M.~Ma, S.~Khanuja, Y.~Zhang \emph{et~al.}, ``{FLEURS}: Few-shot learning evaluation of universal representations of speech,'' \emph{SLT}, 2022.

\bibitem{kreutzer2021quality}
J.~Kreutzer, I.~Caswell, L.~Wang, A.~Wahab \emph{et~al.}, ``Quality at a glance: An audit of web-crawled multilingual datasets,'' \emph{TACL}, 2021.

\bibitem{abdulmumin2022separating}
I.~Abdulmumin, M.~Beukman, J.~O. Alabi, C.~C. Emezue, E.~Asiko, T.~P. Adewumi, S.~H. Muhammad, M.~Adeyemi, O.~Yousuf, S.~Singh, and T.~Gwadabe, ``Separating grains from the chaff: Using data filtering to improve multilingual translation for low-resourced {African} languages,'' \emph{WMÊ}, 2022.

\bibitem{penedo2024fineweb}
G.~Penedo, H.~Kydlíček, L.~B. Allal, A.~Lozhkov, M.~Mitchell \emph{et~al.}, ``{The FineWeb Datasets}: Decanting the web for the finest text data at scale,'' \emph{arXiv preprint arXiv: 2406.17557}, 2024.

\bibitem{jw300}
{\v{Z}}.~Agi{\'c} and I.~Vuli{\'c}, ``{JW}300: A wide-coverage parallel corpus for low-resource languages,'' in \emph{Proceedings of ACL}, A.~Korhonen, D.~Traum, and L.~M{\`a}rquez, Eds.\hskip 1em plus 0.5em minus 0.4em\relax Florence, Italy: Association for Computational Linguistics, jul 2019, pp. 3204--3210.

\bibitem{iroyin}
T.~Ògúnrèmí, K.~Túbosún, A.~Anuoluwapo, I.~Orife, and D.~I. Adelani, ``Ìròyìnspeech: Yorùbá speech corpus,'' \emph{Proceedings of LREC}, 2023.

\bibitem{ibrahim2021}
\BIBentryALTinterwordspacing
U.~A. Ibrahim, ``Hausa speech corpus,'' 2021. [Online]. Available: \url{https://doi.org/10.17632/j6kjmfrbby.2}
\BIBentrySTDinterwordspacing

\bibitem{wanjawa2022kencorpus}
B.~Wanjawa, L.~D.~A. Wanzare, F.~Indede, O.~McOnyango, E.~Ombui, and L.~Muchemi, ``{Kencorpus}: A {Kenyan} language corpus of {Swahili}, {Dholuo} and {Luhya} for natural language processing tasks,'' \emph{Journal for Language Technology and Computational Linguistics}, 2022.

\bibitem{pca}
K.~P. F.R.S., ``{LIII}. {On} lines and planes of closest fit to systems of points in space,'' \emph{Philosophical Magazine Series 1}, vol.~2, pp. 559--572, 1901.

\bibitem{wadasnr}
C.~Kim and R.~M. Stern, ``Robust signal-to-noise ratio estimation based on waveform amplitude distribution analysis,'' in \emph{INTERSPEECH}, 2008, pp. 2598--2601.

\bibitem{soapboxlabs_audio_quality_feature}
\BIBentryALTinterwordspacing
{Soapbox Labs}, ``Audio quality feature,'' 2025, documentation. Accessed: 2025-02-06. [Online]. Available: \url{https://docs.soapboxlabs.com/guides-\&-tutorials/audio-quality-feature/}
\BIBentrySTDinterwordspacing

\bibitem{huggingface}
\BIBentryALTinterwordspacing
{Hugging Face}. (n.d.) Evaluation. Chapter 5 of the Audio Course. [Online]. Available: \url{https://huggingface.co/learn/audio-course/en/chapter5/evaluation}
\BIBentrySTDinterwordspacing

\bibitem{xlsr}
A.~Conneau, A.~Baevski, R.~Collobert, A.~Mohamed, and M.~Auli, ``Unsupervised cross-lingual representation learning for speech recognition,'' in \emph{INTERSPEECH}, 2021, pp. 2426--2430.

\bibitem{seamless2023}
{Seamless Communication}, L.~Barrault, Y.-A. Chung, M.~C. Meglioli, D.~Dale, N.~Dong \emph{et~al.}, ``Seamless: Multilingual expressive and streaming speech translation,'' \emph{ArXiv}, vol. abs/2312.05187, 2023.

\bibitem{ogueji-etal-2021-small}
K.~Ogueji, Y.~Zhu, and J.~Lin, ``{Small Data? No Problem!} {Exploring} the viability of pretrained multilingual language models for low-resourced languages,'' in \emph{Proceedings of the 1st Workshop on Multilingual Representation Learning}, D.~Ataman, A.~Birch, A.~Conneau, O.~Firat, S.~Ruder, and G.~G. Sahin, Eds.\hskip 1em plus 0.5em minus 0.4em\relax Punta Cana, Dominican Republic: Association for Computational Linguistics, nov 2021, pp. 116--126.

\bibitem{whang2021data}
S.~E. Whang, Y.~Roh, H.~Song, and J.-G. Lee, ``Data collection and quality challenges in deep learning: A data-centric ai perspective,'' \emph{arXiv preprint arXiv: 2112.06409}, 2021.

\bibitem{mishra2022proposal}
\BIBentryALTinterwordspacing
S.~Mishra and A.~Arunkumar, ``A proposal to study "{Is} high quality data all we need?",'' \emph{arXiv preprint arXiv: 2203.06404}, 2022. [Online]. Available: \url{https://arxiv.org/abs/2203.06404v1}
\BIBentrySTDinterwordspacing

\bibitem{10764812}
X.~Yu, Z.~Zhang, F.~Niu, X.~Hu, X.~Xia, and J.~Grundy, ``What makes a high-quality training dataset for large language models: A practitioners’ perspective,'' \emph{39th International Conference on Automated Software Engineering (ASE)}, pp. 656--668, 2024.

\end{thebibliography}

\end{document}